# Integrating probabilistic, taxonomic and causal knowledge in abductive diagnosis

**Dekang Lin** and **Randy Goebel**
Department of Computing Science
University of Alberta
Edmonton, Alberta, Canada T6G 2H1
{lindek,goebel}@cs.Ualberta.ca

## Abstract

We propose an abductive diagnosis theory that integrates probabilistic, causal and taxonomic knowledge. Probabilistic knowledge allows us to select the most likely explanation; causal knowledge allows us to make reasonable independence assumptions; taxonomic knowledge allows causation to be modeled at different levels of detail, and allows observations be described in different levels of precision.

Unlike most other approaches where a causal explanation is a hypothesis that one or more causative events occurred, we define an explanation of a set of observations to be an occurrence of a chain of causation events. These causation events constitute a scenario where all the observations are true. We show that the probabilities of the scenarios can be computed from the conditional probabilities of the causation events.

Abductive reasoning is inherently complex even if only modest expressive power is allowed. However, our abduction algorithm is exponential only in the number of observations to be explained, and is polynomial in the size of the knowledge base. This contrasts with many other abduction procedures that are exponential in the size of the knowledge base.

## 1 Introduction

Abduction is the inference to the best explanation. To arrive at the best explanation, an abductive reasoner must make use of information from disparate sources. For the problem of abductive diagnosis, efforts have been made to integrate probabilistic and causal knowledge [Cooper, 1984; Peng and Reggia, 1987; Pearl, 1988; Lin and Goebel, 1990a], causal and taxonomic knowledge [Patil, 1987; Kautz, 1987]. None of these, however, is able to accommodate all three kinds of knowledge.

The difficulties of integrating causal and taxonomic knowledge include the following:

1. Although the links in belief networks [Pearl, 1988; Cooper, 1984] are not limited to causal links, "isa" links cannot be treated in the same ways as other links in the belief networks. (We elaborate in Section 6.2.)

2. Probability alone is not enough to rank explanations. For example, the probability of getting sick is never less than that of getting a cold. However, the former is not necessarily a better diagnosis than the latter.

We propose an abductive diagnosis theory that integrates probabilistic, causal, and taxonomic knowledge. Probabilistic knowledge allows us to select the most likely explanation; causal knowledge allows us to make reasonable independence assumptions; taxonomic knowledge allows causation to be modeled at different levels of detail and observations be described at different levels of precision.

Our model represents domain knowledge with a causal network. Unlike most other approaches, where a causal explanation is a hypothesis that one or more causative events occurred, we define an explanation of a set of observations to be the occurrence of a chain of causation events. These causation events constitute a scenario where all the observations are true. We show that the probabilities of the scenarios can be computed from the conditional probabilities of the causation events.

Computational inefficiency has been a major difficulty with abductive reasoning systems. Abductive reasoning is inherently complex even if only modest expressive power is allowed. Our algorithm, however, is exponential only in the number of observations to be explained, and is polynomial in the size of the knowledge base. This contrasts with many other abduction procedures that are exponential in the size of the knowledge base.

In the next section, we explain the notion of causation event. Section 3 introduces the concept of scenario. In Section 4, we show that the probability of a scenario can be computed from the conditional probability of causation events. In Section 5, the problem of finding the most probable explanation is solved by relating it to the Steiner Problem in Graphs [Dreyfus and Wagner, 1972]. Relationship to other research is discussed in Section 6. Our contributions to abductive diagnosis are summarized in Section 7.

## 2 Causation events

The concept of causation event was first introduced by Peng and Reggia [Peng and Reggia, 1987] to explicitly represent the statement "x actually caused y." A causation event $c \leadsto e$ is true "iff both [the cause event] $c$ and [the effect event] $e$ occur and $e$ is actually being caused by $c$ [Peng and Reggia, 1987, p.149]." One of the motivations for distinguishing causation events from other events in a probabilistic causal world is that $c \leadsto e$ cannot be expressed by a Boolean expression of the events $c$ and $e$, because in situations where both $c$ and $e$ occur, $c \leadsto e$ may still be false.

Unfortunately, Peng and Reggia's definition of causation event does not provide a way to judge whether a causation event occurred, because, unlike other basic events, causation events are usually unobservable. We can only observe the co-occurrence of the cause and effect.

Here we present a definition of causation event following a suggestion from Pearl[1]. First, note that causation can be modeled at different levels of details [Patil, 1987]. A causation event at one level of representation may correspond to a chain of micro causation events at a more refined level. If $c \leadsto e$ corresponds to a chain of micro causation events $c \leadsto i_1, i_1 \leadsto i_2, \ldots i_n \leadsto e$ then, when $c \leadsto e$ occurs, not only must $c$ and $e$ must occur, but also must the intermediate micro events $i_1, \ldots, i_n$.

**Definition 2.1 (causation event)** *A causation event, denoted by $c \leadsto e$, is false iff there do not exist micro events $i_1, \ldots, i_n$ at a more refined level of representation such that $c \leadsto i_1, i_1 \leadsto i_2, \ldots i_n \leadsto e$ is a chain of micro causation events and $e, i_1, \ldots, i_n, c$ co-occur.*

With this definition, $c$ and $e$ must be true if $c \leadsto e$ is true.

## 3 Knowledge representation

We use a language $<E, C, H>$ to represent the domain knowledge, where

1. $E$ is a finite, non-empty set of events, each representing the presence of a disorder, a symptom or a pathological state.

2. $C \subset E \times E$ is the set of causation events.

3. $H \subset E \times E$ is the "isa" relation in $E$. $C$ and $H$ are disjoint. We use $e_1 \xrightarrow{\text{isa}} e_2$ to denote that $e_1$ is an $e_2$ and $\xrightarrow{\text{isa}}*$ to denote the reflexive and transitive closure of the "isa" relation.

The diagrammatic form of the language is a causal network where each node represents an element in $E$. Elements in $C$ and $H$ are represented by *causal* and *"isa"* links respectively. Figure 1 shows an example of causal network.

**Definition 3.1 (scenario)** *A scenario is a pair $(culprit, causations)$, where $culprit \in E$, and $causations \subset C$. Participants in a scenario are the events in $E$ that must be true in the scenario. We define scenario and its participants recursively as follows:*
[1] Personal communication

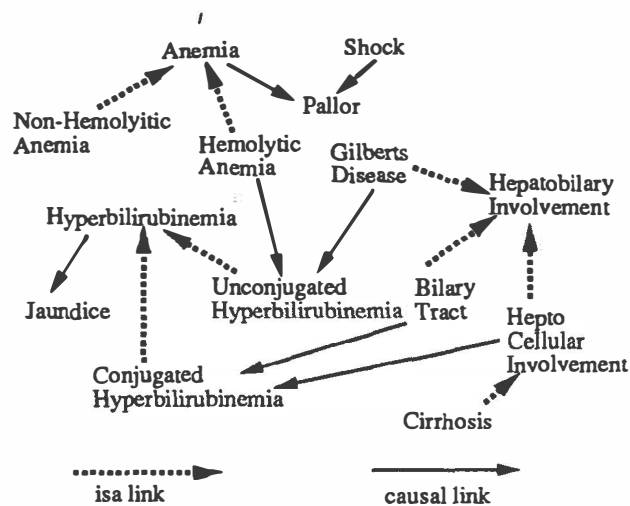

Figure 1: A causal network

1. *Let $c \in E$. Then $(c, \emptyset)$ is a scenario and $participants(c, \emptyset) = \{c\}$.*

2. *Let $(a, \alpha)$ be a scenario. Let $c_0$ be a maximally specific event in $participants(a, \alpha)$, i.e., $\neg \exists d \neq c_0, d \xrightarrow{\text{isa}} *c_0, d \in participants(a, \alpha)$. Let $c_1$ be an event such that $c_0 \xrightarrow{\text{isa}} *c_1$. Let $(c_1, \beta)$ be a scenario such that*

   a. $\alpha \cap \beta = \emptyset$;

   b. $\beta = \{c_1 \leadsto c_2\} \cup \gamma$ and $(c_2, \gamma)$ is a scenario.

   c. *there does not exist a scenario $(c_3, \beta')$ such that $c_0 \xrightarrow{\text{isa}} *c_3 \xrightarrow{\text{isa}} *c_1$, $c_1 \leadsto c_2 \notin \beta'$ and $participants(c_1, \gamma) \cap participants(c_3, \beta') \neq \emptyset$ and $(a, \alpha \cup \beta')$ is a scenario.*

   *Then $(a, \alpha \cup \beta)$ is a scenario and $participants(a, \alpha \cup \beta) = participants(a, \alpha) \cup participants(c_1, \beta)$.*

The condition 2.c ensures that the scenario $(c_1, \beta)$ is not preempted by a more specific scenario when combined with $(a, \alpha)$. When such $(c_3, \beta')$ exists, at least some part of $(c_2, \gamma)$ is related to $c_0$ via a more specific class. $c_1$ is called the *reference class* of $c_0$ with respect to $(c_2, \gamma)$. This notion of reference class is a generalization of the one in [Kyburg Jr., 1982], where the reference class is with respect to a property.

In Figure 2, for example, 2. $(d, \{b \leadsto e, d \leadsto g\})$, 3. $(c, \{a \leadsto e\})$ and 4. $(f, \{f \leadsto g, a \leadsto e\})$ are scenarios; whereas 5. $(d, \{a \leadsto e\})$ is not a scenario, because $a \leadsto e$ is preempted by $b \leadsto e$.

Given a set of observations, explanations are scenarios where the observations are true. Such a scenario can be regarded as a tentative reconstruction of the causal evolution which has led to the observations.

**Definition 3.2 (explanation)** *An explanation for a set of observations $O$ is a scenario $(a, \alpha)$ such that $O \subset participants(a, \alpha)$ and $a \in D$, where $D \subset E$ is the set of disorders.*

For example, in Figure 2, 2. and 4. are explanations for $\{e, g\}$.
41

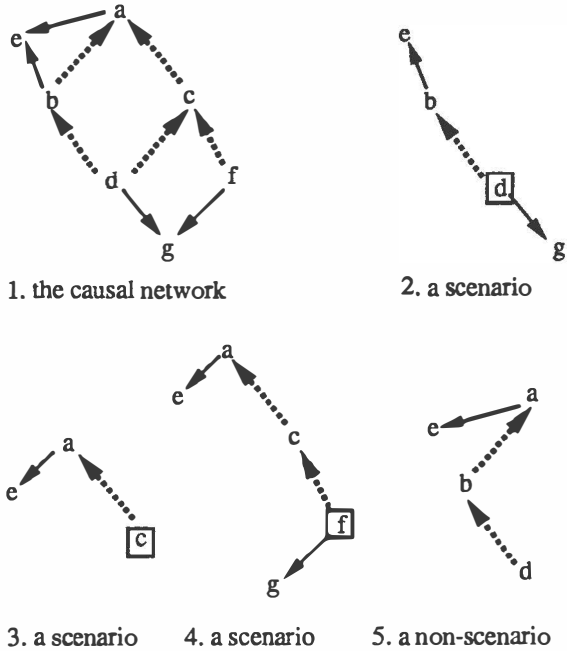

1. the causal network    2. a scenario

3. a scenario    4. a scenario    5. a non-scenario

Figure 2: Scenario are subtrees of the causal network

**Multiple independent disorders** To allow for multiple independent disorders, we can add a distinguished event $\top$ to the set of disorders $D$ and causation events $\top \leadsto d$ for all disorder $d$ such that $d$ has no other causes. $\top$ is always true and $\top \leadsto d$ is true iff $d$ is true. When $\top$ is the culprit of a scenario, the scenario may contain multiple independent disorders.

## 4 Probabilities of scenarios

A probability space is a triple $(\Omega, \mathcal{A}, P)$ [Galambos, 1988], where $\Omega$ is the sample space, $\mathcal{A}$ is a $\sigma$-field of subsets of $\Omega$ and $P$ is a real-valued probability function $P : \mathcal{A} \mapsto [0,1]$.

In the probability space we use, $\Omega$ consists of assignments of truth values to the elements in $E \cup C$, i.e., $\Omega = \{f | f : C \cup H \mapsto \{0,1\}\}$. The set of events, $\mathcal{A}$, consists of all subsets of $\Omega$. Corresponding to each element $e$ of $E \cup C$, there is a basic event in $\mathcal{A}$: $\{f | f \in \Omega, f(e) = 1\}$. Without confusion, we also denote this basic event by $e$. It can be seen that all the events may be constructed by intersection and complement operations on the basic events.

Let $x \in E$, $x \leadsto y \in C$ and $\alpha = \{x_i \leadsto y_i \mid i = 1, \ldots, n, x_i \leadsto y_i \in C\}$, we write $P(x)$, $P(x \leadsto y)$, and $P(\alpha)$ to denote the probabilities of $x$, $x \leadsto y$ and $\alpha$ respectively. A scenario is true if and only if the culprit event is true and all the causation events in the scenario are true. Therefore the probability of a scenario $(a, \alpha)$ is $P(a, \alpha)$, the joint probability of its culprit $a$ and the causation events $\alpha$.

### 4.1 Independence assumption

Let $(a, \alpha)$, $c_0$ and $(c_1, \beta)$ be defined as in Definition 3.1. Then we assume $(a, \alpha)$ is conditionally independent of $\beta$ given $c_1$:
$$P(\beta | c_1, a, \alpha) = P(\beta | c_1)$$

Intuitively, this assumption means that whatever caused $c_1$ can only influence the effects of $c_1$ via $c_1$. This assumption is similar to the axioms of belief networks in [Pearl, 1988].

### 4.2 Probabilities of scenarios

**Theorem 4.1** Let $(a, \alpha)$ be scenario. Then
$$P(a, \alpha) = P(a) \times \prod_{x \leadsto y \in \alpha} P(x \leadsto y | x)$$

**Proof:** This theorem is proved by induction on the structure of scenarios.
*Base Case:* $\alpha = \emptyset$, $P(a, \alpha) = P(a)$.
*Induction Step:* Let $(a, \alpha)$, $c_0$, and $(c_1, \beta)$ be defined as in Definition 3.1. Then, $(a, \alpha \cup \beta)$ is a scenario and

$P(a, \alpha \cup \beta)$
$= P(a, \alpha, \beta)$
$= P(\beta | a, \alpha) \times P(a, \alpha)$
$= P(\beta | c_1, a, \alpha) \times P(a, \alpha)$
   $(\because c_0 \in participants(\alpha)$ and $c_0 \xrightarrow{isa} *c_1)$
$= P(\beta | c_1) \times P(a, \alpha)$
   (by the independence assumption)
$= \frac{P(\beta, c_1)}{P(c_1)} \times P(a, \alpha)$
$= \prod_{x \leadsto y \in \beta} P(x \leadsto y | x) \times P(c_1)/P(c_1)$
   $\times \prod_{x \leadsto y \in \alpha} P(x \leadsto y | x) \times P(a)$
   (by the induction assumption)
$= \prod_{x \leadsto y \in \alpha \cup \beta} P(x \leadsto y | x) \times P(a)$ ∎

**Example 4.2** *The probability of the scenario in Figure 2.2 is*
$$P(d, d \leadsto g, b \leadsto e) = P(d) \times P(d \leadsto g | d) \times P(b \leadsto e | b)$$

**Corollary 4.3** Let $(a, \alpha)$ be a scenario. Then
$$\log(\tfrac{1}{P(a,\alpha)}) = \log(\tfrac{1}{P(a)}) + \sum_{x \leadsto y \in \alpha} \log(\tfrac{1}{P(x \leadsto y | x)})$$

**Estimating the conditional probabilities** In [Lin and Goebel, 1990b], we have shown that under reasonable assumptions about causal influence, $P(e|c) - P(c \leadsto e|c) < P(e)$. Therefore, when $P(e)$ is significantly less than $P(e|c)$, which is usually the case, $P(c \leadsto e|c)$ can be approximated by $P(e|c)$.

## 5 Finding the most probable explanation

We now quantify the causal link from node $x$ to node $y$ by a weight $\log(\tfrac{1}{P(x \leadsto y | x)})$ and associate weight 0 with "isa" links and a weight $\log(\tfrac{1}{P(x)})$ with each node $x$. Then for any explanation $(a, \alpha)$, $\log(\tfrac{1}{P(a,\alpha)})$ is the total weight of all the links in $(a, \alpha)$ plus the weight of the root. Since maximizing $P(a, \alpha)$ is equivalent to minimizing $\log(\tfrac{1}{P(a,\alpha)})$, the problem of finding the most probable explanation becomes that of finding the minimum weight explanation. The latter problem is a variation of a graph-theoretic problem known as the Steiner Problem in Graphs which can be formally stated as follows:

**Definition 5.1 (Steiner Problem in Graphs)** *Let $G = <N, E>$ be a weighted graph, where $N$ is the set of nodes and $E$ is the set of edges. Each edge $e \in E$ is associated with a non-negative weight $w(e)$. Given a set of node $S \subseteq N$, the Steiner Problem in Graphs asks for a sub-tree $T \subseteq E$, such that,*

   a. *all nodes in $S$ are connected together by $T$;*

   b. *$\sum_{e \in T} w(e)$ is minimal.*

*The minimal tree is called the Steiner Tree (connecting $S$).*

It is well known that the Steiner Problem in Graphs is NP-Complete [Garey and Johnson, 1979, p.208]. This, however, does not imply that our model is computationally intractable. The admissible inputs must also be taken into consideration [Levesque, 1989]. A crucial observation here is that the number of observations to be explained in a single case is usually small, much smaller than the number of nodes in the network.

In [Lin and Goebel, 1990a], we present an algorithm for finding the most probable explanations in networks without "isa" links. The algorithm simulates a set of processes distributed over the nodes in the networks. Each process performs the same local algorithm, which consists of receiving, processing and sending messages that are transmitted across the edges. The messages contain optimal solutions for smaller subproblems. These sub-solutions are combined at the receiver node to generate larger optimal solutions, which in turn are sent further. The worst case complexity is $O(3^k n + 2^k e)$, where $k$ is the number of observations to be explained, $n$ and $e$ are the number of nodes and edges in the network, respectively. The average complexity is reduced by exploiting the locality of the nodes to be connected. The nodes that are unrelated to the observations will not be involved in the computation at all.

The algorithm in [Lin and Goebel, 1990a] can be extended to deal with taxonomic structure in the network by adding local constraints to the process at each node. The transmission and combination of the messages are governed by the local constraints such that only the weight of valid scenarios are computed and minimized. The complexity of the revised algorithm is $O(3^k n + k 2^k e)$.

## 6 Relationship to other research

### 6.1 Early expert systems

Research on abductive diagnosis was pioneered by early expert systems such as CASNET [Weiss *et al.*, 1978] and INTERNIST [Miller *et al.*, 1982]. The need for integrating probabilistic, causal and taxonomic knowledge has been recognized in the design of these systems [Finin and Morris, 1988]. Although these systems have achieved significant results, they have often been criticized for their use of poorly defined and unjustified weighting schemes and scoring heuristics as well as unreasonable assumptions about probability distributions.

In contrast, the weighting and scoring scheme in our model are based on formal probability theory. We have also provided a clear specification of the diagnosis task.

### 6.2 Belief Network

Belief networks [Pearl, 1988] are directed acyclic graphs in which each node represents a random variable (or uncertain quantity) which can take on two or more possible values. Causal explanations are instantiations of the variables of the causal network and are obtained by a distributed message propagation. The propagation algorithm, however, is designed for singly connected networks, (i.e. networks with no undirected loops) [Pearl, 1988]. Although he also proposed two extensions to multiply connected networks (clustering and conditioning), both methods are liable to exponential complexity [Henrion, 1987].

Another problem with the belief revision procedure is that an explanation consists of instantiations for all the variables in the network. This implies that every piece of evidence must be propagated to the entire network, even to the totally irrelevant sections of the knowledge base [Pearl, 1988, p.259]. As was discussed in [Lin and Goebel, 1989], our message passing algorithm is able to exploit the locality of the observations to be explained. The nodes that are unrelated to the observations are not activated during the message passing process.

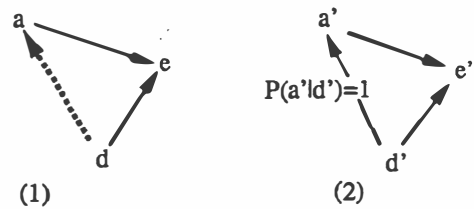

Figure 3: Isa link must be treated differently

"Isa" links cannot be trivially included in belief networks because an $a \stackrel{\text{isa}}{\rightarrow} b$ cannot be simply treated as a link for which $P(b|a) = 1$. Comparing (1) and (2) in Figure 3, for example, the influence of $d$ on $e$ via $a$ should be preempted by the link from $d$ to $e$, whereas in (2), the influence of $d'$ on $e'$ via the two paths should be combined.

### 6.3 Evidential recognition

Shastri [Shastri, 1989] has studied evidential reasoning in a network representation language. Nodes in his network represent concepts, and there are two kinds of links: "isa" links and "has-property" links. A "has-property" link from concept $c$ to the value $v$ of a property $p$ denotes that the concept $c$ may have $v$ as the value for property $p$. The number of instances of $c$ is denoted by $\#c$ and the number of those having $v$ as the value for property $p$ is denoted by $\#c[p,v]$. An example is given in Figure 4. The number in the box is the number of known instances of the concept. The number by an edge is the number of instances having that property value.

An important feature of the representation language is the partial specification of statistical information. If $\#c[p,v]$ is not specified for a concept $c$, it may be inherited from a maximally specific super class $c'$ of $c$ with $\#c'[p, v]$ specified. $c'$ is said to be relevant to $c$ with respect to $[p, v]$.

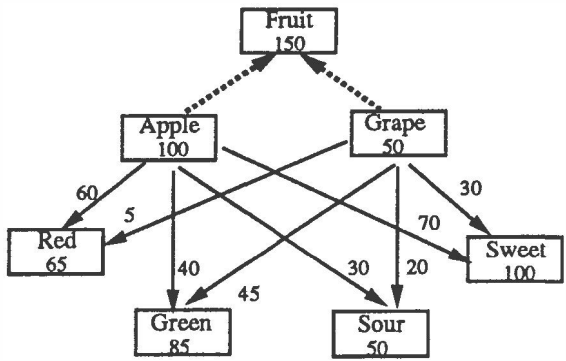

Figure 4: Knowledge about fruits

Evidential recognition may be described as follows: given a set of candidate concepts C-SET= $\{c_1, c_2, \ldots, c_n\}$ and a description DESCR= $\{[p_1, v_1], [p_2, v_2], \ldots, [p_m, v_m]\}$, where $[p_i, v_i]$ is a property value pair, find $c^* \in$ C-SET such that $c^*$ is the most likely concept described by DESCR among C-SET. For example, given C-SET= {apple, grape}, DESCR= {[has-color, green], [has-taste, sour]}, find which of apple and grape is more likely. Evidential recognition is abductive in the sense that one is seeking the concept that best exemplifies the described properties.

Shastri identified two sets of assumptions which, when satisfied, allows the most probable concept to be computed using the principle of maximum entropy. The first is the "Unique relevant concept" assumption [Shastri, 1989, p.337], i.e., for each property-value pair $[p, v]$ in DESCR, there exists a unique concept relevant to a candidate concept $c$ with respect to $[p, v]$. The second assumption is more complex; checking whether a given C-SET satisfies the assumption may take exponential time in the worst case.

The representation language in [Shastri, 1989] is inadequate to represent causal knowledge because unlike properties, which are directly associated with concepts, symptoms may be indirectly caused by disorders. Therefore, Shastri's representation language does not provide any mechanism to model causal chaining. Note further, that the user is required to supply the set of candidate concepts C-SET.

Our algorithm for diagnosis may also be used for evidential recognition in Shastri's network. To do this, we associate $\log(\frac{\#c}{\#c[p,v]})$ with the "has-property" link from concept $c$ to property value $v$ and $\log(\frac{1}{\#c})$ with node $c$. The most probable concept is found by connecting the nodes representing the property values in DESCR, using our algorithm for diagnosis.

When the "Unique relevant concept" assumption is satisfied, $\forall c \in$ C-SET, there is only one scenario $(c, \alpha)$ containing DESCR. The total weight of this scenario is $\log(\frac{1}{\#c}) + \sum_{[p,v] \in \text{DESCR}} \log(\frac{\#c_p}{\#c_p[p,v]})$ where $c_p$ is the relevant concept of $c$ with respect to $[p, v]$. The minimization of the weight is equivalent to maximization of the score in [Shastri, 1989, p.337]: $\#c \prod_{[p,v] \in \text{DESCR}} \frac{\#c_p[p,v]}{\#c_p}$. Therefore, our algorithm gives the same answer as Shastri's.

## 7 Conclusion

We have described a model of abductive diagnosis that is relatively efficient, and that can make use of causal, probabilistic and taxonomic knowledge about a problem domain. Our representation is based on an elaboration of Peng and Reggia's idea of "causation event," extended to account for the intuition behind multiple levels of causation.

The network form of causation relations is similar to Pearl's belief networks, but uses a Steiner Tree algorithm to identify scenarios which constitute the required explanations for observed symptoms. In addition, a natural independence assumption allows relatively efficient computation of the probability of the most likely scenario without requiring that probabilities be propagated throughout the complete network.

Finally, we provide a definition of scenario that accommodates the notion of taxonomic knowledge by providing a restriction to most specific scenarios. The extended definition permits us to express taxonomic relations amongst events, and retain the probabilistic ranking of scenarios.

Much work remains, including the application of our model to provide important feedback about the clarity of the model's semantics, as well as further evidence of its practical efficiency.

## Acknowledgements

This paper builds on earlier work that concentrates on the meaning of causation events and their underlying micro-events. The suggestion for that approach is due to Judea Pearl, who took the time to understand what we were attempting.